\documentclass[10pt,twocolumn,letterpaper]{article}

\usepackage[pagenumbers]{cvpr} 

\usepackage{graphicx}
\usepackage{amsmath}
\usepackage{amssymb}
\usepackage{booktabs}
\usepackage{subcaption} 
\usepackage{tikz}
\usepackage{footmisc}
\usepackage{amsfonts}
\usepackage{mathtools}
\usepackage{array}
\usepackage{tabularx}
\usepackage{textcomp}
\usepackage{pgfplots}
\usepackage{comment}
\usepackage{multirow}
\newcolumntype{Y}{>{\centering\arraybackslash}X}

\definecolor{cvprblue}{rgb}{0.21,0.49,0.74}
\usepackage[pagebackref,breaklinks,colorlinks,citecolor=cvprblue]{hyperref}

\def\paperID{9444} 
\def\confName{CVPR}
\def\confYear{2025}

\title{DATTA: Domain-Adversarial Test-Time Adaptation \\ for Cross-Domain WiFi-Based Human Activity Recognition} 

\author{Julian Strohmayer\qquad Rafael Sterzinger \qquad Matthias Wödlinger\qquad Martin Kampel\\
Computer Vision Lab, TU Wien\\
Favoritenstr. 9/193-1, 1040 Vienna, Austria\\
{\tt\small \{julian.strohmayer, rafael.sterzinger, matthias.woedlinger, martin.kampel\}@tuwien.ac.at}
}

\begin{document}
\maketitle

\begin{abstract}
Cross-domain generalization is an open problem in WiFi-based sensing due to variations in environments, devices, and subjects, causing domain shifts in channel state information. To address this, we propose Domain-Adversarial Test-Time Adaptation (DATTA), a novel framework combining domain-adversarial training (DAT), test-time adaptation (TTA), and weight resetting to facilitate adaptation to unseen target domains and to prevent catastrophic forgetting. DATTA is integrated into a lightweight, flexible architecture optimized for speed.
We conduct a comprehensive evaluation of DATTA, including an ablation study on all key components using publicly available data, and verify its suitability for real-time applications such as human activity recognition. When combining a SotA video-based variant of TTA with WiFi-based DAT and comparing it to DATTA, our method achieves an 8.1\% higher F1-Score. The PyTorch implementation of DATTA is publicly available at: \href{https://github.com/StrohmayerJ/DATTA}{https://github.com/StrohmayerJ/DATTA}.
\end{abstract}

\section{Introduction}
WiFi has emerged as a promising modality in person-centric sensing due to its advantages over optical approaches such as cost-effectiveness, unobtrusiveness, visual privacy protection, and the ability to perform long-range sensing through walls~\cite{YongsenWiFiSurvey2019, Fu234782379}.
Jointly, these qualities enable efficient, contactless monitoring of human activities in confined indoor environments without per-room sensor deployment, providing a significant economic advantage ~\cite{StrohmayerICVS}.

Channel State Information (CSI) serves as the foundation for modern WiFi-based person-centric sensing. CSI is a metric obtained in the Orthogonal Frequency-Division Multiplexing (OFDM) scheme, which subdivides a WiFi channel into multiple sub-channels with different carrier frequencies (subcarriers) \cite{HernandezWiFiOnTheEdgeSurvey}. 
This subdivision allows for fast, parallel transmission of data, while CSI provides detailed information about how each subcarrier is affected by a given domain, enabling the correction of domain-specific noise at the receiver on a per-subcarrier basis.
By correlating the distinctive patterns of amplitude attenuation and phase shifts in CSI caused by specific human activities, tasks such as Human Activity Recognition~(HAR) can be performed ~\cite{Liu45546465465464}.

A significant challenge in WiFi-based HAR that limits practical applications is the poor cross-domain generalization of trained models~\cite{chen2023cross}. In this context, the term \textit{domain} encompasses various factors such as the physical environment, the morphological variability among monitored individuals, the sensing hardware used, and the electromagnetic noise from surrounding WiFi devices.
Altogether, these factors constitute the characteristics of a domain. 
This generalization problem stems fundamentally from the nature of CSI itself: designed to capture amplitude and phase perturbations in a transmitted signal induced by a given domain, models trained on such data are inherently domain-specific and do not generalize well.

While various approaches such as domain-invariant feature extraction and virtual sample generation have been explored to improve cross-domain generalization in WiFi-based HAR~(cf.~\cite{chen2023cross} and references therein), they often have significant limitations: domain-invariant feature extraction typically requires complex and impractical hardware setups, alongside computationally expensive feature pre-processing~\cite{Zhang2022Widar}. On the other hand, virtual sample generation (data augmentation) depends on prior knowledge of the target domain and struggles to adapt to dynamic signal variations over time~\cite{serbetci2023simple}. Given the substantial differences between WiFi domains, models trained exclusively on source domain data often overfit, learning features that are ineffective in new domains. Moreover, WiFi domains can shift unpredictably due to minor environmental changes, such as moving a chair or adding a new WiFi device, which significantly alters signal characteristics. 

For these reasons, an approach is required that not only learns domain-invariant features during training but also adapts to new, unseen domains at test time, while maintaining alignment with the learned domain-invariant feature space to avoid catastrophic forgetting. To address this challenge, we propose Domain-Adversarial Test-Time Adaptation (DATTA) for WiFi-based HAR, a novel framework that combines Domain-Adversarial Training (DAT)~\cite{ganin2016domain} with Test-Time Adaptation (TTA)~\cite{sun2020test}. By learning domain-invariant features during training and continuously adapting to signal variations at inference, DATTA ensures robust performance across diverse and dynamically evolving WiFi domains.

\vspace{-1mm}
\paragraph{Contributions}
To address the significant challenge of cross-domain generalization in WiFi-based HAR, we present the following contributions:

\begin{enumerate}
    \item [(I)] We propose Domain-Adversarial Test-Time Adaptation~(DATTA), a novel approach for WiFi-based HAR that combines domain-adversarial training, test-time adaptation, and targeted regularization to enable robust, real-time adaptation to unseen target domains.
    \item [(II)] We design a lightweight, flexible framework for real-time WiFi-based HAR, optimized for efficient adaptation with DATTA. The code is made publicly available to facilitate further research and reproducibility.
    \item [(III)] We conduct a comprehensive evaluation of DATTA, including an ablation study on all components using publicly available data, and analyze its computational efficiency, demonstrating effectiveness and suitability for real-time applications.
\end{enumerate}

\section{Related Work}
\begin{figure*}[t!]
  \centering
   \includegraphics[width=\linewidth]{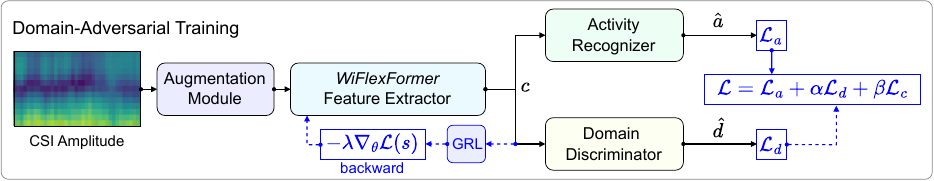}
   \vspace{-6mm}
   \caption{Overview of the domain-adversarial training approach used in DATTA.}
   \label{figDAT}
   \vspace{-2mm}
\end{figure*}

\paragraph{Domain-Adversarial Training}
In their seminal work on domain-adversarial training, Ganin and Lempitsky \cite{ganin2015unsupervised} introduce the Domain-Adversarial Neural Network (DANN). This architecture utilizes a Gradient Reversal Layer (GRL) to align feature representations between source and target domains by minimizing domain discrepancy in an adversarial manner while preserving discriminative task-related features. An extended version of this work \cite{ganin2016domain} provided further theoretical insights and empirical validation, establishing DANN as a cornerstone in domain adaptation research.

Building on DANN, Tzeng et al. \cite{tzeng2017adversarial} improved cross-domain generalization through domain confusion and target-specific classifiers, allowing for more nuanced adaptation to target-specific features. Long et al. \cite{long2018conditional} further advanced DAT with Conditional Domain-Adversarial Networks (CDAN), which integrate conditional task labels to refine alignment between domains and capture complex dependencies.
Peng et al. \cite{peng2019moment} recently extended DAT with Moment Matching for Multi-Source Domain Adaptation (M$^{3}$SDA), which adapts knowledge from multiple labeled source domains to an unlabeled target domain by dynamically aligning higher-order moments of feature distributions. M$^{3}$SDA’s use of multiple classifiers enables domain-specific adaptation but results in higher inference costs, making it less suitable for real-time applications.

In contrast, Jiang et al.~\cite{jiang2018towards} presented one of the first successful applications of DAT in WiFi-based HAR. Their approach, combining a single CNN-based feature extractor with an MLP-based domain discriminator and activity recognizer, maintains lower inference times by using a single classifier.
This architecture is thus better suited for real-time WiFi-based HAR, which is why we combine it with TTA, as explored in this work, to enable effective adaptation to unseen domains while maintaining computational efficiency.

\vspace{-4mm}
\paragraph{Test-Time Training}
Methods on handling distribution shifts at test time can be grouped into two categories~\cite{lin_video_2023}: the first involves modifying the training process, often with a self-supervised auxiliary task; the second approach, more aligned with our method, adapts pre-trained models directly at test time without modifying training.

TTT methods belong to the first group, leveraging auxiliary tasks during training to facilitate adaptation when encountering shifts at test time. Sun et al.~\cite{sun2020test} propose a TTT approach that updates model weights with each test sample, using an architecture with two separate heads, one for self-supervision and another for the main task, enabling real-time adaptation.
Building on this, Gandelsman et al.~\cite{gandelsman2022testtime} employ masked autoencoders and a reconstruction loss to enhance the learning signal.
For video data, Wang et al.~\cite{wang2023testtime} apply TTT with a sliding window of recent frames, enabling continuous updates in response to changing contexts. 
Further advancing TTT, Sun et al.~\cite{sun_learning_2024} propose an approach where a hidden state gets introduced to the model that serves as an adaptive representation, updating with each self-supervised step without preset priors. 

\vspace{-4mm}
\paragraph{Test-Time Adaptation}
In contrast, TTA methods, aligned with the second approach, focus on adapting a pre-trained model during inference without altering its training. Lin et al.~\cite{lin_video_2023} introduce a TTA method for RGB video that adjusts test-time statistics to match those from training, promoting consistency across temporally augmented views to enhance coherence in model predictions.
Wang et al.~\cite{wang_continual_2022} address challenges in continual adaptation with posthoc regularization techniques, averaging predictions across model weights and augmentations to reduce error accumulation and restoring model weights to their original state intermittently to prevent catastrophic forgetting.
Liu et al.~\cite{liu_TTT_nodate} further enhance adaptability by using a self-supervised contrastive learning objective to align source and target features, improving robustness without temporal dependencies. Extending these TTA approaches, Ma et al.~\cite{ma2024improved} propose a so-called improved self-training method that refines pseudo-labels using graph-based correction and stabilizes adaptation with a parameter moving average. 

\section{Domain-Adversarial Test-Time Adaptation}
Our proposed framework, DATTA, combines DAT with TTA to enable robust cross-domain generalization in WiFi-based~HAR. Through DAT, the model learns domain-invariant features by leveraging data from diverse domains, achieving offline adaptation to varied data characteristics. Our DAT architecture builds on the adversarial structure for domain-invariant feature learning outlined in~\cite{jiang2018towards}, with a \textit{WiFlexFormer}-based central feature extractor~\cite{strohmayer2024wiflexformer} to ensure efficient, real-time HAR. To further enhance cross-domain generalization, we incorporate a specialized augmentation module tailored to the unique properties of WiFi CSI. Despite DAT's effectiveness, residual domain shifts can still occur at test time due to environmental changes that lead to distribution shifts in data. To account for these shifts as well, we apply the TTA framework from ~\cite{lin_video_2023} to align the feature distributions of target and source domains, enabling real-time adaptation to previously unseen data distributions during test time. Additionally, we leverage random weight resetting~\cite{vedaldi_adversarial_2020} to prevent catastrophic forgetting of the learned domain invariance, ensuring sustained model stability.

\subsection{Domain-Adversarial Training}
Figure \ref{figDAT} illustrates the DAT architecture used in DATTA, consisting of the \textit{Feature Extractor}, \textit{Activity Recognizer}, and \textit{Domain Discriminator}. The feature extractor processes CSI amplitude data to learn domain-invariant features by generating representations that are informative for activity recognition while disregarding domain-specific aspects. During training, the extracted features are fed to the activity recognizer and domain discriminator. The domain discriminator enforces domain invariance by applying an adversarial loss, pushing the feature extractor to produce features that are indistinguishable across domains. After training, the domain discriminator is discarded, leaving a model optimized for cross-domain generalization.

\vspace{-3mm}
\paragraph{Model Input/Output}
Our DAT architecture takes CSI amplitude spectrograms $s \in \mathcal{S}$ as input, each associated with an activity label $a \in \mathcal{A}$ and a domain label $d \in \mathcal{D}$, where $\mathcal{A}$ and $\mathcal{D}$ represent the sets of activities and domains, respectively. The output is the predicted activity label $\tilde{a} \in \mathcal{A}$.

\vspace{-2mm}
\paragraph{Augmentation Module}
Before feature extraction, raw amplitude spectrograms are processed by the augmentation module, which applies a set of realistic random augmentations to the CSI signal to increase data variability. 
Among these augmentations are amplitude perturbations, circular rotations along the temporal axis, as well as pixel- and row-wise dropout with mean replacement~\cite{StrohmayerICLR}.

\vspace{-3mm}
\paragraph{Feature Extractor}
The feature extractor is based on the \textit{WiFlexFormer}~\cite{strohmayer2024wiflexformer} architecture which is chosen due to its lightweight design and simplicity. It consists of a convolutional stem followed by Gaussian positional encoding and a transformer encoder with class token and a linear classification layer. Its architecture is designed for both, amplitude and Doppler frequency shift features, however, in our experiments, we will exclusively run in the amplitude mode to minimize the parameter count and inference time. The resulting model is comparatively small with only $\approx 40$k parameters.

\vspace{-3mm}
\paragraph{Activity Recognizer}
The activity recognizer, acting as the classification head in our architecture, consists of two linear layers with a ReLU activation function in between and $|\mathcal{A}|$ output channels. It takes the class token embeddings $c$ as input and outputs the predicted activity probabilities~$\hat{a}$.
Its loss $\mathcal{L}_{a}$, the main training objective, consists of the cross-entropy between the predicted and true activities.

\vspace{-2mm}
\begin{equation}
    \vspace{-2mm}
    \mathcal{L}_{a} = -\sum_{k=1}^{|\mathcal{A}|} a_k \log (\hat{a}_k)
    \vspace{-2mm}
\end{equation}

\vspace{-1mm}
\paragraph{Domain Discriminator}
The domain discriminator architecture follows the activity recognizer with two linear layers with a ReLU activation function in between and output channels $|\mathcal{D}|$. It takes the class token embeddings $c$ as input and outputs a prediction over domains. The domain loss $\mathcal{L}_d$ is then computed using the cross-entropy between the predicted domain probabilities $\hat{d}$ and the true domain labels~$d$: 
\begin{equation} 
    \mathcal{L}_d = -\sum_{k=1}^{|\mathcal{D}|} d_k \log (\hat{d}_k).
\end{equation}

Furthermore, to facilitate efficient training without having to freeze model weights alternately, we utilize a GRL. As shown in Figure ~\ref{figDAT}, during the forward pass, the GRL acts as an identity function, allowing the features to flow unchanged to the domain discriminator. However, during backpropagation, it multiplies the gradients $\nabla_{\theta} \mathcal{L}(s)$ by $-\lambda$ to reverse them, thus, returning $-\lambda \nabla_{\theta} \mathcal{L}(s)$ to the feature extractor: 
\vspace{-1mm}
\begin{align} 
    &\text{Forward Pass:} \quad &\operatorname{GRL}(s) &= s, \\
    &\text{Backward Pass:} \quad &\operatorname{GRL}(s) &= -\lambda \nabla_{\theta} \mathcal{L}(s),
\end{align}  
where $\lambda$ is a scaling parameter that controls the strength of the adversarial signal.
By reversing the gradients, the feature extractor is encouraged to produce features that are indistinguishable across domains, thus learning domain-invariant representations.

To perform DAT without overwhelming the feature extractor in the early phase of training, we perform dynamic scaling of $\lambda$ as follows:
\vspace{-3mm}
\begin{equation}
\lambda = \Big(\frac{2}{1 + e^{-10p}} - 1\Big) \gamma,
\vspace{-1mm}
\end{equation}

where $p \in [0,1]$ represents the training progress and $\gamma$ is a scaling parameter to control the adversarial signal strength. This allows the feature extractor to focus on learning robust features for the primary task of activity recognition in the beginning and, by gradually increasing the strength of the adversarial signal, enables a smooth transition to domain-invariant feature learning.

\vspace{-3mm}
\paragraph{Domain-Adversarial Loss}
The loss function in our DAT architecture leverages adversarial training to balance activity recognition and domain-invariant feature learning while penalizing overconfidence in a specific class through a confidence control mechanism. This is achieved by combining task-specific ($\mathcal{L}_{a}$) and domain-specific ($\mathcal{L}_{d}$) losses with the axillary loss $\mathcal{L}_{c}$, representing the \textit{Confidence Control Constraint} (CCC) from \cite{jiang2018towards}. $\mathcal{L}_{c}$ penalizes predictions that are overly certain by adding a penalty for class probabilities approaching 0 or 1. Here, $\hat{a}_{ik}$ represents the predicted probability for activity class $k \in \mathcal{A}$ of the $i$-th sample, ensuring that each class prediction is regularized:
\vspace{-2mm}
\begin{equation}
    \mathcal{L}_{c} = -\sum_{k=1}^{|\mathcal{A}|} \log(\hat{a}_{ik}) + \log(1-\hat{a}_{ik})
\end{equation}

Combined with task- and domain-specific losses, the final domain-adversarial loss function minimized during DAT is given by:
\vspace{-3mm}
\begin{equation}
    \mathcal{L} = \mathcal{L}_{a} + \alpha \mathcal{L}_{d} + \beta \mathcal{L}_{c},
    \vspace{-1mm}
\end{equation}

where $\alpha$ and $\beta$ are weighting parameters, used for controlling the strengths of the adversarial signal and the CCC, respectively. 

\subsection{Test-Time Adaptation}
While DAT is effective in learning domain-invariant features, large domain shifts still lead to a drop in performance during test time.
In order to reduce the impact of such domain shifts, we employ TTA, allowing off-the-shelf pre-trained models to adapt online to new target domains without requiring additional labeled data.

Building upon the framework proposed by Lin et al.~\cite{lin_video_2023}, we adapt TTA from RGB video to CSI amplitude spectrograms to further enhance the generalization of DAT during test time by performing feature distribution alignment, i.e., aligning source statistics of the model with online estimates of the target statistics.
Additionally, to prevent overfitting to the target distribution during prolonged adaptation, i.e. catastrophic forgetting~\cite{vedaldi_adversarial_2020}, we implement random weight resetting, following the approach proposed by Wang et al.~\cite{wang_continual_2022}. 
Specifically, a subset of model parameters is reverted to their source model values to keep them closer to the domain-invariant feature space.

\vspace{-2mm}
\paragraph{Feature Map Alignment}
To address the distribution shift, we align the statistics of feature maps, i.e., matching the means and variances, computed for both the training and test spectrograms.
Let $\phi_l(s; \theta)$ represent the feature map of the $l$-th layer of network $\phi$, computed for a spectrogram $s$ with parameters $\theta$.
Each feature map is a matrix of dimensions $(t_l, f_l)$, where $t_l$ and $f_l$ correspond to the time steps and frequency channels (subcarrier), respectively.

Computing the mean of the $l$-th layer features for a dataset~$\mathcal{S}$ across the time dimension results in a mean vector of size $f_l$, which can be expressed as:

\begin{equation}
\mu_l(\mathcal{S}; \theta) = \mathbb{E}_{s \in \mathcal{S}} \mathbb{E}_{t \in [1,t_l]}\Big[\phi_l(x; \theta)[t]\Big],
\end{equation}
and the variance of the $l$-th layer features is given by:

\begin{equation}
\sigma_l^2(\mathcal{S}; \theta) = \mathbb{E}_{s \in \mathcal{S}} \mathbb{E}_{t\in [1,t_l]} \Big[ (\phi_l(x; \theta)[t] - \mu_l(\mathcal{S}; \theta))^2 \Big].
\end{equation}
For the remainder of this work, we denote the mean and variance computed on the training set with $\Bar{\mu}_l$ and $\Bar{\sigma}_l^2$.
When training data is unavailable, these statistics can be estimated from batch norm layers as well, though with a small decrease in performance~\cite{lin_video_2023}.

At test time, updates are performed iteratively, adjusting the discrepancy between the test statistics of a batch~$\mathcal{B}$ of selected layers $L$ with those computed during training:

\begin{equation}
\mathcal{L}_{\text{TTA}} = \sum_{l \in L} ||\mu_l(\mathcal{B}; \theta) - \bar{\mu}_l||_2 + ||\sigma_l^2(\mathcal{B}; \theta) - \bar{\sigma}_l^2||_2.\label{eq:align_loss}
\end{equation}
In our experiments, we observe that optimal results are obtained by selecting $L$ to contain only the first out of four transformer encoder layers that compose the \textit{WiFlexFormer}-Encoder.

Given the low inference time of the \textit{WiFlexFormer}, we perform TTA for the most realistic application scenario: online, on data received in a stream.
Hence, we chose $|\mathcal{B}|=1$ and continuously evaluate target statistics using exponential moving averages instead of repeatedly computing statistics for the constantly growing test set. 
In other words, given the spectrogram $s_i$, received in iteration $i$, we update the mean and variance estimates as follows:
\begin{align}
&\hat{\mu}_l^{(i)}=\alpha \cdot \mu_l(s_i; \theta) + (1 - \alpha) \cdot \hat{\mu}_l^{(i-1)},\\
&\hat{\sigma}_l^{2(i)}= \alpha \cdot \sigma_l^2(s_i; \theta) + (1 - \alpha) \cdot \hat{\sigma}_l^{2(i-1)},
\end{align}
where $1-\alpha$ denotes the momentum.
As a starting point, we select the source statistics, i.e., $\mu_l^{(0)}=\Bar{\mu}_l$ and $\sigma_l^{2(0)}=\Bar{\sigma}_l^2$. 
Without modifications to Eq.~\ref{eq:align_loss} and with no extensive recomputation necessary, we compute $\mathcal{L}_{\text{TTA}}$ using these estimates instead.

\vspace{-2mm}
\paragraph{Weight Resetting}
In order to avoid catastrophic forgetting, in each iteration, we reset a subset of the current weights to their original values from $\bar{\theta}$, the source models's parameters.
Specifically, consider the weights of layer $l$ in iteration $i$, denoted as~$\theta^{(i)}_l$. 
To randomly reset these, we define a Boolean mask $m_l$ with $\text{dim}(\theta^{(i)}_l) = \text{dim}(m_l)$, where each element of the mask is sampled from a Bernoulli distribution with reset rate $p$. The updated parameter vector~$\theta^{(i)}_l$ is then:

\begin{equation}
\theta^{(i)}_l = m_l \odot \bar{\theta}_l + (1 - m_l) \odot \theta^{(i)}_l,
\end{equation}
where $\odot$ denotes element-wise multiplication.

\vspace{-2mm}
\paragraph{Comparison to TTT and Other TTA Methods}
Unlike TTT, TTA requires no changes to the architecture, making it suitable for off-the-shelf models.
Additionally, inference with TTA is arguably faster than with TTT, as TTT often involves a computationally intensive reconstruction task as the secondary objective \cite{sun_learning_2024, gandelsman_test-time_2022}.
Compared to other TTA approaches that align features by adjusting only the running statistics of normalization layers~\cite{mirza_norm_2022, ma2024improved}, our method updates the entire parameter vector $\theta$ up to the highest layer in~$L$, offering greater flexibility during adaptation compared to the others.
However, this can also be problematic since continuously adapting entire parameter vectors will unlearn learned domain invariant feature transformation eventually.
To overcome this, weight resetting is employed, allowing the model to recover its original form.
As evidenced in Section~\ref{sectionCrossDomainAdapation}, while TTA does increase inference time compared to the original \textit{WiFlexFormer}, practical deployment in real-time applications remains feasible.

\vspace{-1mm}
\section{Evaluation}
\vspace{-1mm}
We evaluate DATTA’s effectiveness through a detailed ablation study of its components, including the augmentation module, loss function elements, and discriminator input in DAT, as well as random weight resetting in TTA to assess its impact on model stability. Our experiments use a modified version of a publicly available dataset adapted for DAT and TTA. We also analyze inference time to assess the computational impact of TTA and weight resetting on real-time performance, addressing practical deployment feasibility.

\begin{figure}[t!]
  \centering
   \includegraphics[width=0.7\linewidth]{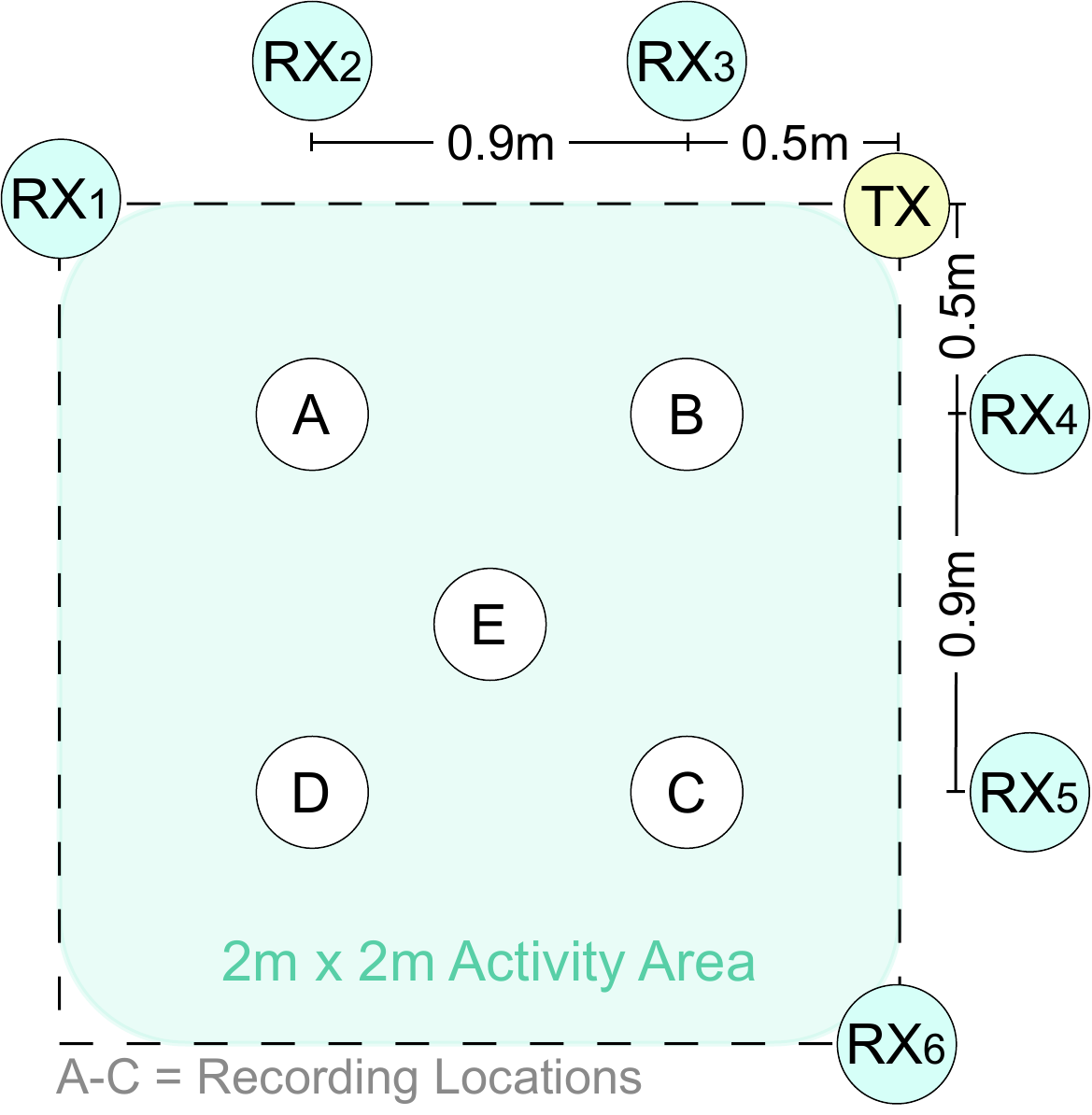}
   \vspace{-2mm}
   \caption{Widar3.0 recording setup featuring a single transmitter~TX and six receivers RX$_{1-6}$. A-E indicate the positions at which activities (hand gestures) are carried out.}
   \label{figWidar3Setup}
   \vspace{-4mm}
\end{figure}

\subsection{Data}
\paragraph{Widar3.0-G6 Dataset} The Widar3.0 dataset \cite{Zhang2022Widar} is one of the most widely used WiFi HAR datasets, featuring CSI recordings of 22 human hand gestures performed by 16 participants in three different indoor environments. Since not all 22 gestures are consistently performed in all three environments, a subset of Widar3.0, referred to as Widar3.0-G6~\cite{Hou2024RFBoostUA}, is often utilized instead of the full dataset. This subset includes 6 hand gestures (\textit{push \& pull}, \textit{sweep}, \textit{clap}, \textit{slide}, \textit{draw circle} and \textit{draw zigzag}) that are performed across all three environments by 16 users, resulting in a total of 68,246 hand gesture samples. The recording setup, shown in Figure \ref{figWidar3Setup}, consists of one 5.825 GHz WiFi transmitter (TX) and six receivers (RX$_{n}$), each equipped with an \textit{Intel WiFi Link 5300} wireless NIC that has three antennas. The CSI of 90 subcarriers (3 antennas $\times$ 30 subcarriers) is collected at each receiver using the Linux CSI Tool~\cite{Halperin43578934578}, utilizing a packet sending rate of 1,000 Hz.

\vspace{-2mm}
\paragraph{Widar3.0-G6D Dataset} 
To evaluate cross-domain generalization, we preprocess the Widar3.0-G6 dataset by extracting CSI from 30 subcarriers of the first antenna at each receiver and applying temporal sub-sampling to 100 Hz for improved computational efficiency. Only samples between 120 and 220 WiFi packets (1.2–2.2 seconds) are retained, with shorter samples zero-padded to standardize length. Amplitude features are then extracted and normalized using min-max scaling. The resulting Widar3.0-G6D dataset comprises 58,648 samples and is split into two disjoint subsets based on unique room-participant combinations (domains): a training subset with 7 domains (2 environments, 7 persons) and a test subset with 9 domains (1 environment, 9 persons), as shown in Table~\ref{tableDataset}. 

\subsection{Model Training}
We evaluate four model configurations to assess the impact of DATTA and its components: the baseline \textit{WiFlexFormer}~($W$), its TTA variant~($W_{\text{TTA}}$), the DAT model~($W_{\text{DAT}}$), and the final DATTA model~($W_{\text{DATTA}}$). All model backbones are trained on \textit{Train}, validated on \textit{Val}, with TTA hyperparameters tuned on \textit{Val}$_{\text{TTA}}$, and tested on the shuffled dataset \textit{Test}, if not stated otherwise~(cf. Table~\ref{tableDataset}).
Note that shuffling \textit{Test} poses an extreme case of high-frequency domain shifts. A Python script that creates these exact subsets from Widar3.0-G6 is provided for reproducibility.

The baseline $W$ uses the vanilla \textit{WiFlexFormer} architecture without DAT or TTA. $W_{\text{DAT}}$ incorporates DAT with loss weights $\alpha=0.3$, $\beta=0.2$, and GRL-scaling $\gamma=8$. Adding TTA to $W$ and $W_{\text{DAT}}$ produces $W_{\text{TTA}}$ and $W_{\text{DATTA}}$, allowing the model’s first layer to adapt during inference. The complete DATTA framework, $W_{\text{DATTA}}$, is tuned using Bayesian and grid search. For all models using TTA, random weight resetting with $p=1\times10^{-4}$ is evaluated to enhance stability. We further assess our augmentation module’s effectiveness by training each model with and without augmentations. Detailed hyperparameters for all model configurations are provided in the supplementary material.

\begin{table}[t!]
\centering
\begin{tabularx}{0.475\textwidth}{>{\arraybackslash}p{9mm}>{\centering\arraybackslash}p{11mm}>{\centering\arraybackslash}p{6mm}>{\centering\arraybackslash}p{9mm}>{\centering\arraybackslash}p{11mm}>{\raggedleft\arraybackslash}p{12mm}}
\toprule
Subset & Activities & Envs. & Persons & Domains & Samples\\
\midrule
\text{Train} & 6 & 2 & 7 & 7 & 19,586\\
\text{Val} & 6 & 2 & 7 & 7 & 4,896\\
\midrule
$\text{Val}_{\text{TTA}}$ & 6 & 1 & 9 & 9 & 3,417\\
\text{Test} & 6 & 1 & 9 & 9 & 30,749\\
\midrule
Total & 6 & 3 & 16 & 16 & 58,648\\
\bottomrule
\end{tabularx}
\vspace{-2mm}
\caption{Overview of the distribution of activities, environments, participants, domains, and activity samples across the Widar3.0-G6D subsets used for training, validation, and testing.}
\label{tableDataset}
\vspace{-4mm}
\end{table}

\begin{table}[t!]
\centering
\begin{tabularx}{0.475\textwidth}{>{\arraybackslash}p{18mm}>{\centering\arraybackslash}p{5mm}>{\centering\arraybackslash}p{5mm}>{\centering\arraybackslash}p{18mm}>{\centering\arraybackslash}p{18mm}}
\toprule
Model & A & R & ACC $\uparrow$ & F1-Score $\uparrow$ \\
\midrule
\multirow{2}{*}{$W$~(baseline)} & - &   & 38.69 \scriptsize{$\pm$ 3.17} & 40.62 \scriptsize{$\pm$ 2.93}\\ 
& \checkmark &   & 47.75 \scriptsize{$\pm$ 4.31} & 49.32 \scriptsize{$\pm$ 4.45}\\ 
\midrule
\multirow{2}{*}{$W_{\text{TTA}}$}  & -&   - & 42.76 \scriptsize{$\pm$ 4.21} & 44.89 \scriptsize{$\pm$ 3.93}\\ 
& \checkmark& -  & 51.70 \scriptsize{$\pm$ 4.40} & 53.20 \scriptsize{$\pm$ 4.26}\\
\midrule
\multirow{2}{*}{$W_{\text{DAT}}$} & - &  & 39.54 \scriptsize{$\pm$ 2.80} & 42.02 \scriptsize{$\pm$ 2.62}\\ 
& \checkmark &     & 64.53 \scriptsize{$\pm$ 1.87}& 65.66 \scriptsize{$\pm$ 1.85}\\ 
\midrule
\multirow{3}{*}{$W_{\text{DATTA}}$}  & - &  -  & 45.26 \scriptsize{$\pm$ 5.83} & 48.23 \scriptsize{$\pm$ 5.11}\\ 
& \checkmark&  - & 65.90 \scriptsize{$\pm$ 1.53} & 67.29 \scriptsize{$\pm$ 1.43}\\
& \checkmark & \checkmark  & \underline{66.92} \scriptsize{$\pm$ 1.54} & \underline{68.13} \scriptsize{$\pm$ 1.45}\\ 
\bottomrule
\end{tabularx}
\vspace{-2mm}
\caption{Cross-domain HAR performance on the Widar3.0-G6D dataset averaged over three runs, comparing models trained with conventional training ($W$), test-time adaptation ($W_{\text{TTA}}$), domain-adversarial training ($W_{\text{DAT}}$), and the proposed combined approach, domain-adversarial test-time adaptation ($W_{\text{DATTA}}$). Columns $A$ and $R$ indicate whether data augmentation and random weight resetting (during TTA) are applied.}
\label{tableResultsWidar3g6D}
\vspace{-4mm}
\end{table}

\subsection{Results}
\paragraph{Augmentation Module} 
We begin our evaluation with the augmentation module, the first component in our DAT pipeline.
Inspecting Table~\ref{tableResultsWidar3g6D}, which depicts all cross-domain activity recognition results, reveals the critical role of our data augmentation approach for handling cross-domain variations.
Comparing the baseline model $W$ with and without augmentation, we observe a substantial improvement in F1-Score, from 40.62\% to 49.32\%.

Its impact is even more pronounced in $W_{\text{DAT}}$, the DAT model:
without augmentation, $W_{\text{DAT}}$ achieves an F1-Score of only 42.02\%, reflecting poor generalization and significant overfitting to domain-specific features.
Hence, augmentation is crucial for DAT as without it, the model overfits early and fails to learn domain-invariant features at later stages of training when the adversarial signal strength is increased.
However, with augmentation, $W_{\text{DAT}}$ attains an F1-Score of 65.66\%, showing that sufficient data variability is essential to facilitating domain-invariant feature learning.

\vspace{-3mm}
\paragraph{Confidence Control Constraint} 
Next, we consider the impact of the CCC which is designed to prevent overconfidence in specific class predictions to encourage a more balanced feature representation across classes.
By discouraging extreme confidence in particular classes, CCC helps stabilize training, making the model more adaptable to unseen domains. 
By means of a hyperparameter search, we identified the optimal weight $\beta=0.2$ for CCC and evaluate its effectiveness, we conducted an ablation study comparing models trained with and without CCC (i.e., $\beta=0$). 
Observing the results, shown in Table \ref{tableAblationCCC}, shows that including a CCC improves performance in both the $W_{\text{DAT}}$ and $W_{\text{DATTA}}$ models.
For $W_{\text{DAT}}$, enabling CCC leads to an increase in F1-Score from 63.81\% to 65.66\%, and for $W_{\text{DATTA}}$, from 65.62\% to 68.13\%, reinforcing that CCC plays a valuable role in enhancing cross-domain generalization.

\begin{table}[t!]
\centering
\begin{tabularx}{0.475\textwidth}{>{\arraybackslash}p{16mm}>{\centering\arraybackslash}p{16mm}>{\centering\arraybackslash}p{18mm}>{\centering\arraybackslash}p{18mm}}
\toprule
Model & CCC & ACC $\uparrow$ & F1-Score $\uparrow$ \\
\midrule
$W_{\text{DAT}}$ &-   & 62.81 \scriptsize{$\pm$ 1.09} & 63.81 \scriptsize{$\pm$ 1.09}\\ %
$W_{\text{DAT}}$ & \checkmark  & \underline{64.53} \scriptsize{$\pm$ 1.87} & \underline{65.66} \scriptsize{$\pm$ 1.85}\\ %
\midrule
$W_{\text{DATTA}}$ & -  & 65.62 \scriptsize{$\pm$ 1.08} & 66.78 \scriptsize{$\pm$ 1.08}\\ %
$W_{\text{DATTA}}$ & \checkmark & \underline{66.92} \scriptsize{$\pm$ 1.54} & \underline{68.13} \scriptsize{$\pm$ 1.45}\\ %
\bottomrule
\end{tabularx}
\vspace{-2mm}
\caption{Ablation study on the CCC, weighted with $\beta=0.2$.}
\label{tableAblationCCC}
\vspace{-5mm}
\end{table}

\vspace{-3mm}
\paragraph{Random Weight Resetting} 
Following this, we evaluate the impact of random weight resetting during TTA which has been introduced to prevent catastrophic forgetting.
Our results, presented in Table~\ref{tableResultsWidar3g6D}, indicate that random weight resetting promotes cross-domain generalization, however only when the model is domain-invariant to some degree, as achieved through DAT with data augmentation.

For $W_{\text{DATTA}}$ with augmentation, enabling random weight resetting boosts the F1-Score to 68.13\%, up from 67.29\% without resetting, achieving the highest performance within our model landscape.
However, in other cases, it has limited or even negative effects. For instance, applying weight resetting to $W_{\text{DATTA}}$ without augmentation or to $W_{\text{TTA}}$ leads to reduced performance.
We hypothesize that this is due to the weights of the base model $W$ and $W_{\text{DAT}}$ without augmentation not being sufficiently domain-invariant yet, causing weight resets to revert beneficial adaptations during TTA.
In contrast, $W_{\text{DATTA}}$, trained with augmentation and thus exhibiting stronger domain invariance, allows weight resetting to maintain proximity to this invariant space during TTA.
Consequently, it prevents drifting due to over-adaptation to specific test domains, avoiding catastrophic forgetting and stabilizing performance across diverse domains.

\begin{figure}[t!]
  \centering
   \includegraphics[width=1\linewidth]{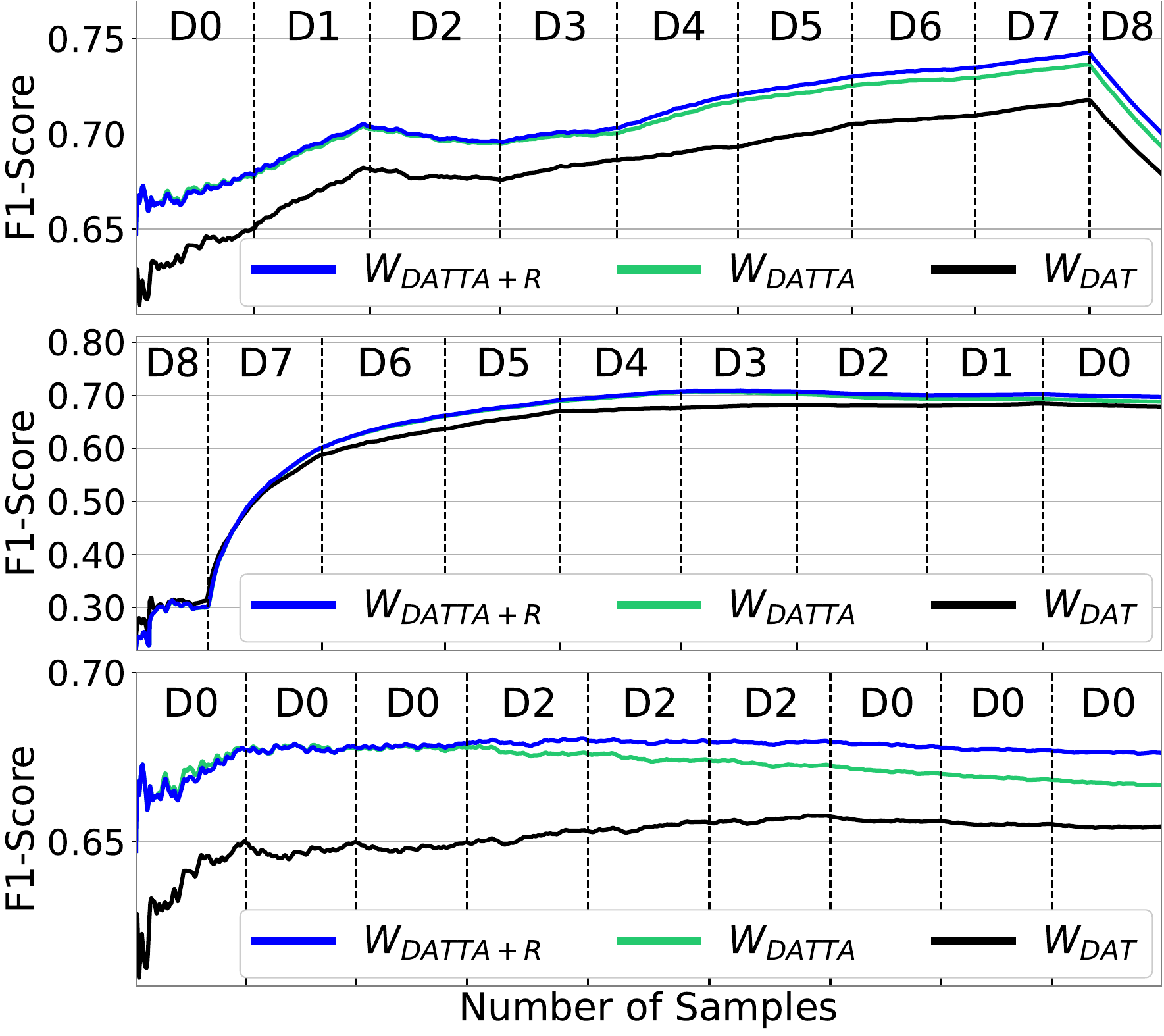}
   \vspace{-5mm}
   \caption{TTA performance across continuous test domain sequences. From top to bottom: (1) ascending domain order (D0 to D8), (2) descending domain order (D8 to D0), and (3) alternating domain order with prolonged domains D0 and D2. Depicted are F1-Scores computed with a rolling window of 100 samples for DATTA models with weight resetting ($W_{\text{DATTA+R}}$, blue), without weight resetting ($W_{\text{DATTA}}$, green), and the baseline DAT model without TTA ($W_{\text{DAT}}$, black).}   
   \label{domainSwitch}
   \vspace{-5mm}
\end{figure}

\vspace{-3mm}
\paragraph{Cross-Domain Adaptation} \label{sectionCrossDomainAdapation}
Moving forward, Figure \ref{domainSwitch} illustrates the performance of three models: DATTA with weight resetting ($W_{\text{DATTA+R}}$), DATTA without weight resetting ($W_{\text{DATTA}}$), and DAT without TTA as the baseline~($W_{\text{DAT}}$), across three domain sequences to evaluate adaptability and resilience to domain shifts.

In the first experiment (top plot), where domains are processed one after another (D0 to D8) to simulate smooth and realistic domain shifts, TTA consistently improves in performance across domains, showing effective incremental adaptation except for the last domain, D8, where it experiences a significant drop.
On closer analysis, we find that D8 has a highly skewed class distribution (with class 1 containing only 8 samples, approximately 0.3\% of the total samples in D8) and substantially different data statistics compared to the other domains.
While these factors may hinder adaptability, we hypothesize that mislabeled activities are the primary cause of poor performance, potentially compounded by the steady, linear decline in accuracy, which suggests systematic misclassification. However, due to the lack of ground-truth image data, we cannot confirm this hypothesis.
Nevertheless, TTA adapts well on all other domains, indicating that participant-induced domain gaps can be overcome effectively.

In the second experiment (middle plot), we process the domains in reversed order (D8 to D0) to test the ability to recover from domains with stark differences in underlying data statistics.
After finishing adaptation to the ill-posed domain, D8, TTA quickly recovers and exceeds baseline performance, demonstrating its resilience in recovering from challenging domains.

Finally, the third experiment (bottom plot) evaluates resilience to catastrophic forgetting during prolonged adaptation to a single domain by alternating between domains D0 and D2.
Here, $W_{\text{DATTA+R}}$ maintains stable performance across domain shifts, demonstrating its robustness in retaining learned features across shifts.
In contrast, $W_{\text{DATTA}}$ continually decreases in performance, indicating a gradual loss of learned domain invariance, which suggests that weight resetting effectively prevents catastrophic forgetting.

Overall, both $W_{\text{DATTA+R}}$ and $W_{\text{DATTA}}$ significantly outperform the baseline $W_{\text{DAT}}$, confirming TTA’s effectiveness in enhancing cross-domain generalization. Additionally, $W_{\text{DATTA+R}}$ outperforms $W_{\text{DATTA}}$ across all experiments, validating the role of random weight resetting in preventing performance degradation under repeated, especially prolonged domain shifts.

\begin{table}[t!]
\centering
\begin{tabularx}{0.475\textwidth}{>{\arraybackslash}p{13mm}>{\arraybackslash}p{22mm}>{\centering\arraybackslash}p{16mm}>{\centering\arraybackslash}p{16mm}}
\toprule
\multicolumn{2}{l}{Model} & ACC $\uparrow$ & F1-Score $\uparrow$ \\

\midrule
$W$&+ViTTA~\cite{lin_video_2023}  & 47.76 \scriptsize{$\pm$ 4.33} & 49.54 \scriptsize{$\pm$ 4.43}\\ 
$W$&+DAT~\cite{jiang2018towards} & 61.87 \scriptsize{$\pm$ 1.39} & 63.16 \scriptsize{$\pm$ 1.55}\\ 
$W_{\text{TTA}}$&+DAT~\cite{jiang2018towards} & 64.48 \scriptsize{$\pm$ 1.58} & 65.87 \scriptsize{$\pm$ 1.41}\\ 
$W_{\text{DAT}}$&+ViTTA~\cite{lin_video_2023} & 64.37 \scriptsize{$\pm$ 2.05} & 65.60 \scriptsize{$\pm$ 1.94}\\ 
\midrule
$W$ & +both & 61.66 \scriptsize{$\pm$ 0.81} & 63.04 \scriptsize{$\pm$ 0.92}\\ 
\multicolumn{2}{l}{$W_{\text{DATTA}}$} & \underline{66.92} \scriptsize{$\pm$ 1.54} & \underline{68.13} \scriptsize{$\pm$ 1.45}\\ %
\bottomrule
\end{tabularx}
\vspace{-2mm}
\caption{SotA comparison against DAT~\cite{jiang2018towards} and ViTTA~\cite{lin_video_2023}. ($W$ and $W_{\text{TTA}}$) + DAT use the discriminator input $c \oplus \hat{a}$, ($W$ and $W_{\text{TTA}}$) + ViTTA, employ the $\ell_1$-loss, and initialize targets with zero, with slightly tuned hyperparameters. Note that all models are trained with augmentation, and $W_{\text{DATTA}}$ additionally uses weight resetting.}
\label{tableSota}
\vspace{-5mm}
\end{table}

\vspace{-3mm}
\paragraph{SotA Comparison}
To compare the proposed DATTA, with existing SotA methods, we evaluate it against the default DAT model as proposed in~\cite{jiang2018towards} and the video-based TTA variant in~\cite{lin_video_2023}. To the best of our knowledge, we are the first to port this video-based TTA variant to the WiFi domain, which presents challenges for direct comparison. To align as closely as possible with the original work, we retain all parameters, tuning only the most crucial ones: the layers for statistic alignment (choosing the first layer instead of the last two) and the learning rate ($1 \times 10^{-6}$).

Shown in Table~\ref{tableSota}, are the results of our SotA evaluation; note that all variants have been trained using the suggested data augmentation module.
When combining ViTTA~\cite{lin_video_2023} with $W$ or $W_\text{DAT}$, the F1-Score stays almost the same ($+0.4$\%/$-0.1$\%), indicating that employing the method as-is has no benefit.
Consequently, evaluating it against our TTA variants, namely $W_\text{TTA}$ and $W_\text{DATTA}$ without weight resetting, performance is worse (49.54\% versus 52.2\% and 67.29\%), suggesting that adopting ViTTA to the WiFi domain is not straightforward, requiring changes in the loss function, target initialization, and extensive parameter tuning. 

Similarly, when combining DAT~\cite{jiang2018towards} with $W$ or $W_\text{TTA}$, performance is significantly improved over $W$, though remains below our DAT variants (63.16\% vs. 65.66\% and 67.29\%), i.e., $W_\text{DAT}$ and $W_\text{DATTA}$.
Here, the key difference is the input to the discriminator: class token embeddings~$c$ concatenated with (\cite{jiang2018towards}) and without (ours) activity logits~$\hat{a}$.
We hypothesize that this is due to~$\hat{a}$ inherently including domain information, namely priors over activities, e.g., lying being more likely performed in the bedroom than in the office. Hence, by including $\hat{a}$, the activity recognizer is penalized for learning these priors, making predictions less accurate.

In a final evaluation, we compare combining both SotA methods naively ($W$+both) against our method $W_\text{DATTA}$: as before, the impact of ViTTA is negligible, resulting in an improvement over the SotA by 8.1\%.

\vspace{-4mm}
\paragraph{Inference Time}
Table \ref{tableInferenceTime} compares the inference times for the baseline model $W$, $W_{\text{DAT}}$, $W_{\text{TTA}}$, and $W_{\text{DATTA}}$ on both an Nvidia RTX 2070 GPU and a Jetson Orin Nano single-board computer. On the RTX 2070, $W$ achieves the fastest inference time at 1.98 ms per sample, with a minor increase to 2.05 ms for $W_{\text{DAT}}$ due to the DAT component. Models incorporating TTA show a notable rise in inference time due to additional adaptation steps, with $W_{\text{TTA}}$ reaching 9.86 ms and $W_{\text{DATTA}}$ reaching 10.05 ms, a ~5x increase over the baseline. With random weight resetting, inference time increases further to approximately 19.52 ms for $W_{\text{TTA}}$ and 20.21 ms for $W_{\text{DATTA}}$, roughly a 10x increase.

Switching from the RTX 2070 to the Jetson Orin Nano incurs an additional ~5x increase in inference time across models, with $W_{\text{DATTA}}$ achieving around 115.39 ms per sample with weight resetting. However, this latency remains sufficient for HAR, achieving approximately 9 frames per second. Although weight resetting adds to inference time, it significantly improves model stability by preventing catastrophic forgetting, which is crucial for sustained performance across domain shifts. Notably, our TTA code is not optimized for speed, and further optimizations could reduce adaptation time.

\begin{table}[t!]
\centering
\begin{tabularx}{0.475\textwidth}{
    >{\arraybackslash}p{10mm}   
    >{\centering\arraybackslash}p{11mm} 
    >{\centering\arraybackslash}p{3mm} 
    >{\raggedleft\arraybackslash}p{16mm} 
    >{\raggedleft\arraybackslash}p{23mm} 
}
\toprule
\multicolumn{3}{c}{} & \multicolumn{2}{c}{Inference Time [ms]} \\ 
\cmidrule(lr){4-5}
Model & Params. & R & \small{RTX 2070} & \small{Jetson Orin Nano}\\ 
\midrule
$W$ & 40.80 k & & \underline{1.98} \scriptsize{$\pm$ 0.21} & \underline{9.24} \scriptsize{$\pm$ 0.25}\\
$W_\text{DAT}$ & 41.97 k & & 2.05 \scriptsize{$\pm$ 0.23} & 10.30 \scriptsize{$\pm$ 0.18}\\
\midrule
\multirow{2}{*} {$W_{\text{TTA}}$} & 40.80 k & - & 9.86 \scriptsize{$\pm$ 1.42} & 56.73 \scriptsize{$\pm$ 1.29}\\
& 40.80 k & \checkmark & 19.52 \scriptsize{$\pm$ 2.27} & 111.39 \scriptsize{$\pm$ 3.93}\\
\midrule
\multirow{2}{*} {$W_{\text{DATTA}}$} & 41.97 k & - & 10.05 \scriptsize{$\pm$ 1.40} & 57.63 \scriptsize{$\pm$ 1.31}\\
& 41.97 k & \checkmark & 20.21 \scriptsize{$\pm$ 2.05} & 115.39 \scriptsize{$\pm$ 3.76}\\
\bottomrule
\end{tabularx}
\vspace{-2mm}
\caption{Inference time comparison for the base model $W$, TTA model $W_{\text{TTA}}$, DAT model $W_{\text{DAT}}$, and DATTA model $W_{\text{DATTA}}$. Column R indicates the use of random weight resetting. Mean inference time is reported over 1,000 iterations (after 100 warm-up iterations) with batch size 1 on an Nvidia RTX 2070 GPU and a Jetson Orin Nano single-board computer.}
\label{tableInferenceTime}
\vspace{-5mm}
\end{table}
\vspace{-1mm}
\section{Conclusion}
\vspace{-1mm}
This work addressed the challenge of cross-domain generalization in WiFi-based HAR by introducing DATTA, a framework combining DAT, TTA, and regularization techniques to enable real-time adaptation to unseen domains. DATTA was integrated into a lightweight, efficient \textit{WiFlexFormer}-based architecture. Through a comprehensive ablation study, we evaluated a component of DATTA -- including the augmentation module, loss functions, discriminator input, and random weight resetting -- and confirmed their effectiveness in enhancing cross-domain generalization and robustness against catastrophic forgetting. Our results showed that DATTA not only outperforms models relying on either DAT or TTA but also adapts reliably to domain shifts, recovers quickly from performance drops, and achieves inference speeds that meet real-time requirements. We provide a PyTorch implementation of DATTA and dataset tools to support future work in adaptive WiFi-based HAR and cross-domain generalization.

{
    \small
    \bibliographystyle{ieeenat_fullname}
    \bibliography{main}
}

\end{document}